\documentclass[letterpaper, 10 pt, journal, twoside]{IEEEtran}

\ifCLASSINFOpdf
\usepackage[pdftex]{graphicx}
\else
\usepackage[dvips]{graphicx}
\fi

\usepackage{amsmath} \usepackage{amsfonts,amssymb,bm}  
\usepackage{xcolor}
\usepackage{tikz}
\usetikzlibrary{fit,positioning}
\usepackage[capitalise]{cleveref}
\usepackage{subcaption} \usepackage{siunitx}
\usepackage[hyphens]{url}
\usepackage{nicefrac}
\usepackage{algorithm}
\usepackage[noend]{algpseudocode}
\usepackage{multirow}
\usepackage{tabularx}
\usepackage{siunitx}
\usepackage{cite}
\crefname{section}{Sec.}{Sec.}

\makeatletter
\renewcommand{\ALG@beginalgorithmic}{\small}
\makeatother

\newcommand{\obsf}{y}

\newcommand{\dom}{\mathcal{X}}

\newcommand{\inx}{x}
\newcommand{\inxopt}{x_{*}}

\newcommand{\stdnoise}{\sigma_\text{no}}
\newcommand{\varnoise}{\stdnoise^2}
\newcommand{\thres}{c}

\newcommand{\thresopt}{\hat{\thres}}

\newcommand{\ls}{+1}
\newcommand{\lu}{0}
\newcommand{\fs}{\bm{g}_\text{s}}
\newcommand{\fu}{\bm{g}_\text{u}}

\newcommand{\f}{\bm{g}}
\newcommand{\Y}{\bm{y}}

\newcommand{\sobs}{\mathcal{D}}
\newcommand{\sobst}[1]{\mathcal{D}_n^{#1}}

\newcommand{\GP}[2]{\mathcal{GP}\left( #1, #2\right)}
\newcommand{\GPCR}[2]{\mathcal{GPCR}\left( #1, #2\right)}

\newcommand{\Ys}{\Y_\text{s}}

\newcommand{\Ns}{N_\text{s}}

\newcommand{\N}[2]{\mathcal{N}( #1 ; #2 )}

\newcommand{\mt}{\tilde{\bm{m}}}
\newcommand{\vart}{\tilde{\Sigma}}

\newcommand{\Esub}[2]{\mathbb{E}_{#1}\left[ #2 \right]}

\newcommand{\Prob}[1]{\text{Pr}(#1)}

\newcommand{\prob}[1]{p( #1 )}

\newcommand{\df}{\text{d}\f}

\newcommand{\CDF}[1]{\Phi\left( #1 \right)}

\newcommand{\R}{\mathbb{R}}

\newcommand{\meanpost}{\bm{\mu}_\text{EP}}
\newcommand{\varpost}{\bm{\Sigma}_\text{EP}}

\newcommand{\gpcr}{{\sc GPCR}}
\newcommand{\eicc}{{\sc Eic$^2$}}

\newcommand{\pibu}{{\sc \small PIBU}}
\newcommand{\safeopt}{{\sc \small SafeOpt}}

\DeclareMathOperator*{\argmaxaux}{argmax}
\newcommand{\argmax}{\displaystyle\argmaxaux}
\DeclareMathOperator*{\argminaux}{argmin}
\newcommand{\argmin}{\displaystyle\argminaux}

\begin{document}

\title{Robot Learning with Crash Constraints}

\author{Alonso Marco$^{1}$, Dominik Baumann$^{4,1}$, Majid Khadiv$^{1}$, Philipp Hennig$^{2,1}$, \\ Ludovic Righetti$^{3,1}$ and Sebastian Trimpe$^{4,1}$\thanks{Manuscript received: October, 15, 2020; Revised December, 8, 2020; Accepted January, 25, 2021.}\thanks{This paper was recommended for publication by Editor D. Kuli\'{c} upon evaluation of the Associate Editor and Reviewers' comments.
This work was supported by the International Max Planck Research School for Intelligent Systems (IMPRS-IS).} \thanks{$^{1}$Max Planck Institute for Intelligent Systems, T\"{u}bingen/Stuttgart, Germany;
		{\tt\footnotesize \{amarco,dbaumann,mkhadiv\}@tuebingen.mpg.de}
		$^{2}$University of T\"{u}bingen, Computer Science Department, 72074 T\"{u}bingen, Germany;
		{\tt\footnotesize philipp.hennig@uni-tuebingen.de}
		$^{3}$New York University, Brooklyn, NY 11201 USA;
		{\tt\footnotesize ludovic.righetti@nyu.edu}
		$^{4}$Institute for Data Science in Mechanical Engineering, RWTH Aachen University, 52068 Aachen, Germany;
		{\tt\footnotesize trimpe@dsme.rwth-aachen.de}
		}\thanks{Digital Object Identifier (DOI): see top of this page.}
}
\markboth{IEEE Robotics and Automation Letters. Preprint Version. Accepted January, 2021}
{Marco \MakeLowercase{\textit{et al.}}: Robot Learning with Crash Constraints}

\maketitle

\begin{abstract}

In the past decade, numerous machine learning algorithms have been shown to successfully learn optimal policies to control real robotic systems.
However, it is common to encounter failing behaviors as the learning loop progresses. Specifically, in robot applications where failing is undesired but not catastrophic, many algorithms struggle with leveraging data obtained from failures. This is usually caused by (i) the failed experiment ending prematurely, or (ii) the acquired data being scarce or corrupted. Both complicate the design of proper reward functions to penalize failures. In this paper, we propose a framework that addresses those issues. We consider failing behaviors as those that violate a constraint and address the problem of \emph{learning with crash constraints}, where no data is obtained upon constraint violation. The no-data case is addressed by a novel GP model (GPCR) for the constraint
that combines discrete events (failure/success) with continuous observations (only obtained upon success).
 We demonstrate the effectiveness of our framework on simulated benchmarks and on a real jumping quadruped, where the constraint threshold is unknown a priori.
Experimental data is collected, by means of constrained Bayesian optimization, directly on the real robot.
Our results outperform manual tuning and GPCR proves useful on estimating the constraint threshold.

\end{abstract}

\IEEEpeerreviewmaketitle

\section{Introduction}

\IEEEPARstart{D}{uring} the past decades, developments in machine learning (ML) have boosted the usage of algorithms that learn directly from data in real systems. 
 In the context of robotics, data-efficient reinforcement learning methods have allowed to increase automation and mitigate human intervention during the learning process. An increasingly popular family of algorithms that has made this possible is
Bayesian optimization (BO) \cite{shahriari2016taking}, which has been proposed to iteratively search for optimal controller parameters from a few evaluations. To compensate for data scarcity, prior assumptions are placed on the unknown performance objective using a Gaussian process (GP) \cite{Rasmussen2006Gaussian} model. This is iteratively optimized by collecting informative data points through experiments.

While BO has shown impressive results in learning from data on real robots, such as bipedal locomotion \cite{calandra2016bayesian,rai2018bayesian,yeganegi2019efficient,yuan2019bayesian}, quadrotor hovering \cite{berkenkamp2016safe}, and manipulation \cite{marco2016automatic,englert2018learning,driess2017constrained}, not all issues have been solved yet.
 For example,
during the search, some controller parametrizations may yield unstable behavior (i.e., one of the robot states growing unboundedly).
This can lead to the robot failing to accomplish the task, for instance, by falling down \cite{calandra2016bayesian,rai2018bayesian,antonova2016sample,yeganegi2019efficient}, or by having to emergency-stop it prematurely \cite{marco2016automatic,berkenkamp2016safe}.
While failures must be completely avoided in safety-critical applications \cite{berkenkamp2016safe}, herein we consider scenarios where failing is undesired but not catastrophic. In such scenarios, failures still provide a valuable source of information, yet, designing effective reward functions to account for failures is difficult. Indeed, failures are
often handled by using heuristic user-defined fixed penalties to the reward. Such practice usually requires expert (or domain) knowledge \cite{yuan2019bayesian,marco2016automatic,englert2018learning,driess2017constrained,calandra2016bayesian,rai2018bayesian,antonova2016sample}.
For example, quantifying the reward of a quadruped that falls down while learning to jump implies analyzing the state trajectories and calibrating a substantially different reward from that of successful jumps. This requires detailed knowledge about the platform and the task itself.

\begin{figure}[!t]
\centering
\includegraphics[width=\columnwidth]{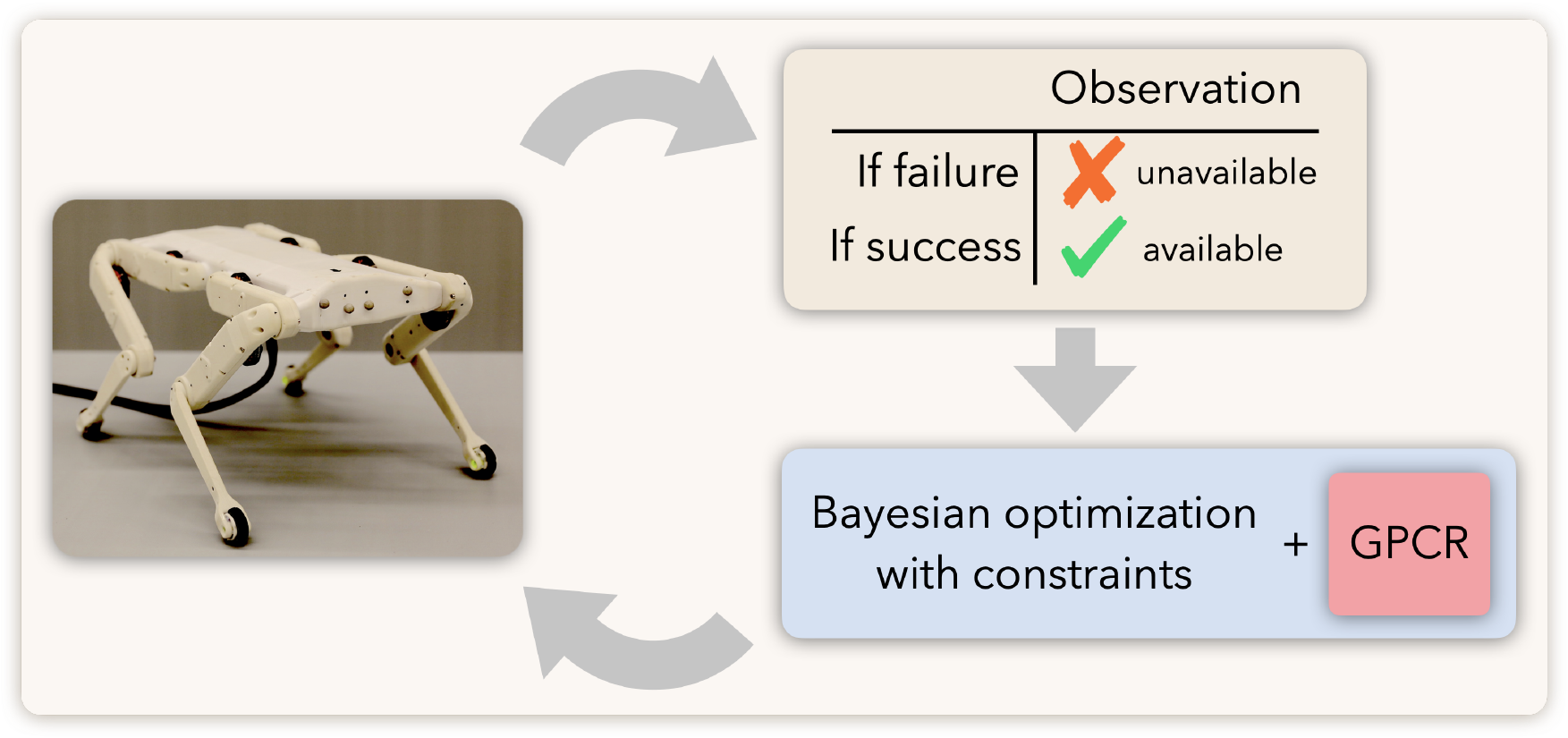}
\caption{Learning loop with a quadruped that learns to jump using Bayesian optimization with constraints and GPCR. 
While we acquire height and current measurements for successful jumps, failed jumps report no observations.
}
\label{fig:eye_catcher}
\end{figure}

Because expert knowledge might not always be available (or impractical to expect it), we propose herein a reward function that is \emph{not defined} at failing parametrizations, i.e., those in which the robot fails to complete the task. In this paper, we refer to this as the problem of \emph{learning with crash constraints} (LCC): we obtain a reward value in case of successful execution, but in case of a failure we only know that the robot failed. We
address it using Bayesian optimization with unknown constraints (BOC).

In a nutshell, BOC extends BO by adding black-box constraint functions that are expensive-to-evaluate, modeled with GPs. Such framework avoids visiting areas where the constraints are violated, as these are revealed during the search, hence speeding up convergence to the constrained optimum \cite{griffiths2017constrained,snoek2013bayesian,gelbart2014bayesian,schonlau1998global}.

Because in BOC it is assumed that both, the constraint and the reward function are defined upon failure, it cannot be used straightforwardly to tackle the 
LCC problem.
Depending on the nature of the constraint, BOC assumes that it can report either a real number \cite{hernandez2016general,schonlau1998global,gelbart2014bayesian} or a binary outcome \cite{lindberg2015optimization,snoek2013bayesian,bachoc2019gaussian} when evaluated.
In the former case, the constraint is usually modeled with standard GP regression, while in the latter it is modeled with a GP classifier \cite[Sec. 3.2]{Rasmussen2006Gaussian}. 
However, in the 
LCC problem,
none of those two modeling options are adequate. On one hand, a GP regressor is impractical because no real values are reported upon failure. On the other hand, a GP classifier (GPC) expects binary observations while, upon success, the constraint observations are real valued. In such case, modeling the constraint using GPC implies discarding valuable information about ``how well was the constraint satisfied'' that could otherwise be leveraged throughout the optimization process to avoid visiting unpromising regions.

Herein, we propose a novel GP model for the constraint that integrates \emph{hybrid} observations, i.e., (i) a discrete label indicating whether the task was accomplished or not and (ii) a continuously valued observation only in case the task was accomplished. Due to the conceptual closeness of this model with a GP classifier, but reinterpreting the observations in the GP regressor, we call it \emph{Gaussian process for classified regression} (\gpcr). 
As opposed to existing multi-output Gaussian process models,
\gpcr~combines discrete labels and continuous observations in a single-output GP model. 

The \gpcr~model is tailored to the 
LCC problem
and exhibits two main benefits with respect to existing approaches for robot learning with BO.
 First, it removes the need of penalizing failure cases with user-defined penalties to the reward. Instead, those failure cases are captured in the \gpcr~that models the constraint.
Second, \gpcr~treats the constraint threshold as a hyperparameter and estimates it from data.
This brings two advantages along: (i) mitigating the need of expert knowledge to define the constraint threshold, which is usually a challenging problem \cite{berkenkamp2016safe}, and (ii) reducing possible human bias introduced by ad hoc choices.

\textbf{Contributions:}
This paper proposes three main contributions. 
First, a novel BOC framework that specifically addresses the 
LCC problem
by (i) transferring the modeling effort of robot failures to the constraint and (ii) mitigating failures during learning. Within this framework, any available BOC acquisition function can be used. However, because the popular expected improvement with constraints (EIC) \cite{gelbart2014bayesian} is employed herein, we name our framework \emph{expected improvement with crash constraints} (\eicc).
In general, \eicc~becomes useful when the ``reason for failure'' can be measured upon success, but estimating the constraint threshold requires large amounts of expert knowledge.
Second, we propose a novel single-output GP model (\gpcr) for the constraint that handles hybrid data (discrete labels and real values) and treats the constraint threshold as a hyperparameter, hence reducing the need for expert knowledge. Third, we propose for the first time a solution
to the 
LCC problem
when learning from data on a real system, specifically, on a real jumping quadruped robot (cf. \cref{fig:eye_catcher}). Our approach enabled the quadruped to learn significantly higher jumps than those found with manual tuning.

\textbf{Related work:}
In contexts other than robotics, the 
LCC problem
first appeared in \cite{donskoi1993partially}, and has adopted different names ever since \cite{antonio2019sequential,bachoc2019gaussian}.
In the field of global optimization, a number of recent BOC algorithms have been proposed to address the 
LCC problem
on simulated scenarios \cite{lindberg2015optimization,lee2011optimization}, and numerical benchmarks \cite{antonio2019sequential}.

In the area of robot learning with BO, the 
LCC problem
is present, but
it has not been treated so far with well founded methodologies.
Instead, 
it is usually overcome with ad hoc heuristic strategies that range in diversity depending on the task at hand.
A common heuristic is to discard any acquired data upon failure and assign a user-defined pre-fixed penalty to the reward. This approach was followed
in \cite{yuan2019bayesian} to prevent the torso of the robot from rotating above a certain angle, and in \cite{marco2016automatic} to prevent some robot states from leaving a pre-defined safety zone.
Another heuristic is to use whatever data becomes available before failing. For example, in \cite{calandra2016bayesian,rai2018bayesian,antonova2016sample}, 
the reward of a walking robot is the distance walked before falling, which acts as a penalty for the fall.

As in this work, EI has previously been extended to address the 
LCC problem,
where the constraint is modeled using different forms of a GP classifier, such as logistic regression \cite{lindberg2015optimization} and probit regression  \cite{snoek2013bayesian}.
Also in the context of robotics, \cite{englert2018learning,driess2017constrained} propose a different BOC algorithm that specifically addresses the 
LCC problem.
Therein, the constraint is modeled as a GP classifier, and its observations are assumed to be binary. 
Such references differ from this work
in two main aspects: (i) contrary to \gpcr, the GP that they use to model the constraint does not explicitly account for hybrid data (discrete labels and continuous values), (ii) the constraint 
threshold is inherently assumed zero or fixed, while we treat it as a learnable hyperparameter, and (iii) their method expands only locally the initial safe region.

In \cite{pourmohamad2016multivariate,zhang2019bayesian,ru2019bayesian}, several multi-output Gaussian process models are proposed to tackle continuous and discrete observations. They jointly model the objective as a GP regressor and a set of binary constraints as GP classifiers. Albeit the 
LCC problem
can also be addressed with such models, (i) they are restricted to cases in which observations are given as binary, (ii) they cannot learn the constraint threshold from data, and (iii) the applicability of such models to robot settings is unclear.

																		 \section{Preliminaries}

\subsection{Problem setting: robot learning with crash constraints}
\label{sec:controlprob}
Let a dynamical system be controlled with a parametrized policy to execute a specific task.
A real experiment successfully executed on the system with policy parameters $x \in \dom_\text{S} \subseteq \dom \subset \R^D$ yields a performance $f(x)$, where the mapping $f:\dom_\text{S} \rightarrow \R$ is unknown. The goal is to solve the constrained optimization problem
\begin{equation}
\inxopt = \argmin_{x\in\dom_\text{S}} f(x),
\label{eq:min_cost_cons}
\end{equation}
where $\dom_\text{S}$ is a safe region, outside of which the controller parameters make the system \emph{crash} during task execution. Such safe region is determined as $\mathcal{X}_\text{S} = \{x : g_1(x) \leq c_1, \ldots, g_G(x) \leq c_G\}$, where $g_j:\dom_\text{S} \rightarrow \R,\;\forall j=1,\ldots,G$ are $G$ 
constraints that indicate whether the aforementioned task succeeds or not and $c_j$ is the constraint threshold.
We assume coupled evaluations of $f$ and $g_j$, i.e., all queries $f(x), g_1(x), \dots, g_G(x)$ are obtained simultaneously by running a robot experiment with controller parametrization $x$.
Furthermore, neither the objective $f$ nor the constraints $g_j$ are defined outside the safe region $\mathcal{X}_\text{S}$. 
Finally, the constraint thresholds $c_j$ are unknown.

\subsection{Gaussian process (GP)}
\label{ssec:GPs}
We model the objective as a Gaussian process, $f \sim \GP{0}{k(x,x^\prime)}$, with covariance function $k : \dom \times \dom \rightarrow \R$, and zero prior mean. Observations $\obsf(x) = f(x) + \varepsilon$
are modeled using additive Gaussian noise $\varepsilon \sim \N{\varepsilon}{0,\varnoise}$.
After having collected $n$ observations from the objective $\sobst{} = \{{\bm{x}}_n , {\bm{y}}_n \} = \{ {x}_1,\dots,{x}_n,{y}_1,\dots,{y}_n \}$, its predictive distribution at location $x$ is given by $\prob{f|\sobst{},x} = \N{f(x)}{\mu(x|\sobst{}),\sigma^2(x|\sobst{})}$, with predictive mean $\mu(x|\sobst{}) = \bm{k}^\top_n(x)[K_n + \varnoise I ]^{-1}{\bm{y}}_n$, where the entries of the vector $\bm{k}_n(x)$ are $[\bm{k}_n(x)]_i = k(x_i,x)$, the entries of the Gram matrix $K_n$ are $[K_n]_{i,j} = k({x}_i,{x}_j)$, and the entries of the vector of observations ${\bm{y}}_n$ are $[{\bm{y}}_n]_i = {y}_i$
. The predictive variance is given by $\sigma^2(x|\sobst{}) = k(x,x) - \bm{k}_n^\top(x)[K_n + \varnoise I ]^{-1}\bm{k}_n(x)$.
In the remainder of the paper, we drop the dependency on the current data set $\sobst{}$ and write $\mu(x)$, $\sigma(x)$ to refer to $\mu(x|\sobst{})$, $\sigma(x|\sobst{})$, respectively.

\subsection{Bayesian optimization with and without constraints}
\label{ssec:boc}
We address the constrained optimization problem \eqref{eq:min_cost_cons} using BOC, where both the objective $f$ and each constraint $g_i$ are modeled with a GP.
The GP posterior is used by BOC algorithms to steer the search toward regions inside $\dom$ where the constraints are satisfied with high probability. To this end, 
the next candidate point $x_{n+1}$
is iteratively computed as maximizer of an acquisition function $\alpha : \dom \rightarrow \R$, i.e., $x_{n+1} = \arg\max_{x\in \dom} \alpha(x)$.
The acquisition $\alpha$ implicitly depends on the GP models of the objective $f$ and the constraints $g_i$ conditioned on the acquired data points. 

In this work we use \emph{expected improvement with constraints} (EIC) \cite{gelbart2014bayesian}. EIC extends the well-known \emph{expected improvement} (EI) \cite{mockus1978toward}, which is a BO strategy for unconstrained problems, to the constrained case.
In the following, we describe both methods.

\subsubsection{Expected improvement (EI)}
\label{sssec:ei}
This algorithm 
reveals areas of low cost function values that are expected to improve upon the
best observation so far. The improvement is defined as $\max \{ \eta - f(x) , 0 \}$ and its expectation is given by
\begin{equation*}
\alpha_\text{EI}(x) = \Esub{f(x) \sim \prob{f|\sobst{},x}}{\max \{ \eta - f(x) , 0 \}},
\label{eq:ei}
\end{equation*}
where $\eta = \min_{r \in \{1,\ldots,n\}} y_r$ is the best observation so far.

\subsubsection{Expected improvement with constraints (EIC)}
\label{sssec:eic}
Akin to EI, this method defines the improvement as $\max \{ \eta_\text{cons} - f(x) , 0 \}$, where $\eta_\text{cons} = \min_{r \in \{1,\ldots,n\}} y_r,\;\text{s.t. } g_j(x_r) \leq 0 \;\forall j$ is the best constrained observation so far. As EI, it reveals areas where such improvement is high in expectation.
In addition, it requires those areas to be safe with high probability. Its acquisition function is given by 
\begin{equation}
\hfilneg
\alpha_\text{EIC}(x) = \Esub{f(x) \sim \prob{f|\sobst{},x}}{\max \{ \eta_\text{cons} - f(x) , 0 \}} \Gamma (x),
\hspace{1000pt minus 1fil}
\label{eq:eic}
\end{equation}
where $\Gamma (x) = \prod_{j=1}^G \Prob{g_j(x) \leq 0}$ and $\Prob{g_j(x) \leq 0}$ is the probability of satisfying constraint $j$.
Although \eqref{eq:eic} requires knowing $\eta_\text{cons}$, such value can only be computed when at least one safe data point has been found. In the absence of safe data points (e.g., at early stages of the exploration), EIC ignores the objective $f$ and explores using the information of the constraint only, i.e., $\alpha_\text{EIC}(x) = \Gamma (x)$.

 \section{Bayesian optimization with crash constraints}

In this section, we introduce the \eicc~framework, which
differs from standard BOC in two aspects, described next.

\begin{figure}[!t]
\centering
\includegraphics[width=\columnwidth]{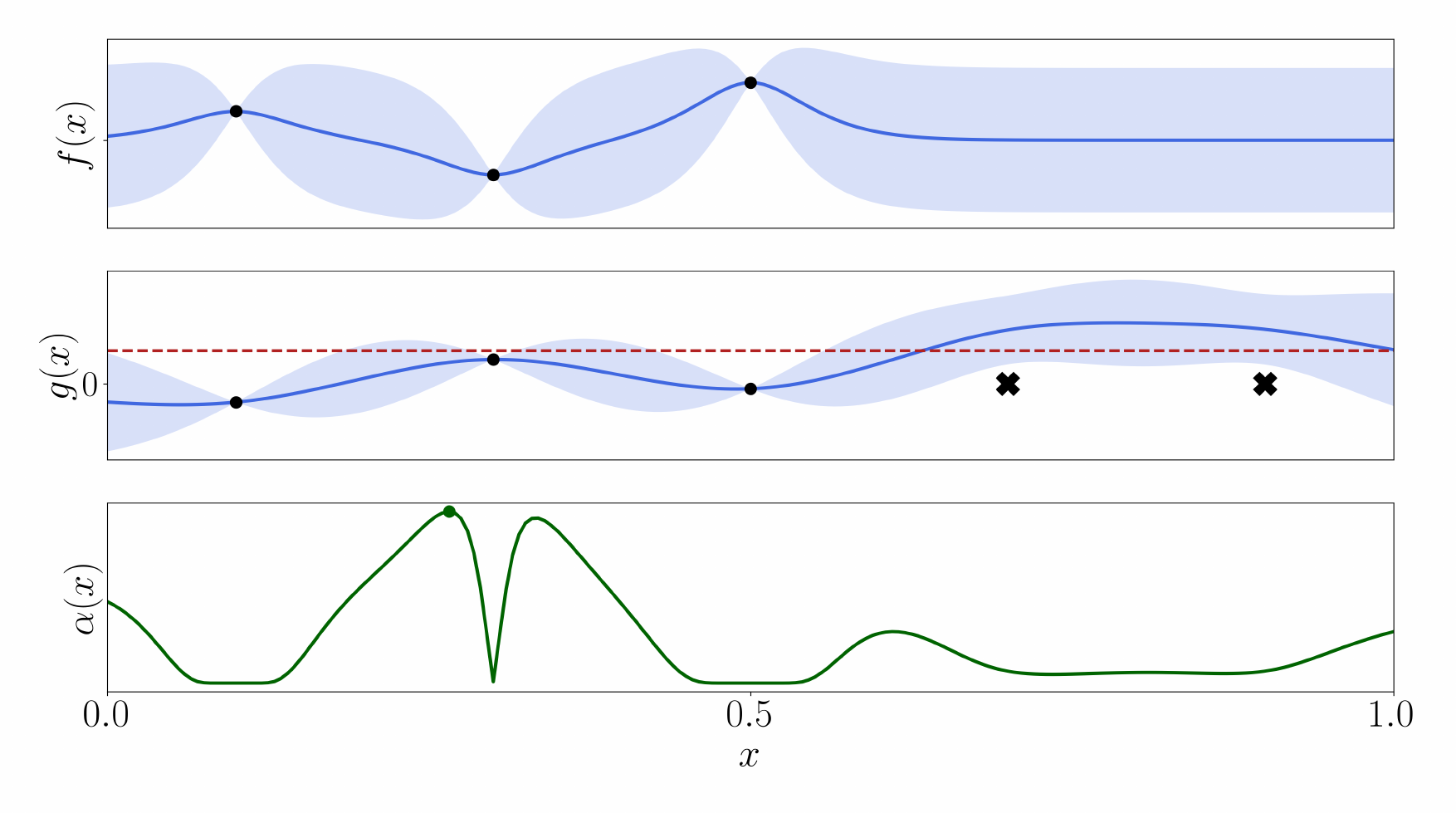}
\caption{Top: Standard GP that models the unknown objective $f$, conditioned on three successful observations. Predictive mean (solid line) and $\pm 2$ standard deviation (surface path). Middle: GPCR model of the unknown constraint $g$, conditioned on five observations: three successful (circles) and two failures (crosses, both plotted at zero for convenience). The GPCR model pushes the probability mass above the learned constraint threshold $\thresopt$ (dashed line) near the two failure observations.
Bottom: EIC acquisition function and next candidate point (green dot) that maximizes it.}
\label{fig:GPCR_all}
\end{figure}

The first difference is about the way each constraint $g_j$ is modeled. While BOC typically uses either a GP regressor or a GP classifier,
\eicc~tackles the absence of data by using a novel GP model: \gpcr. Although we explain in detail GPCR in \cref{sec:gpcrsec}, herein we provide a high-level intuition. \gpcr~is updated with a discrete label indicating ``failure'' when the constraints are violated and with a real value when the constraint is satisfied. The GPCR model is illustrated on a one-dimensional toy example in \cref{fig:GPCR_all} (middle). Therein, the two failures (crosses) are given as discrete labels; they push upwards the probability mass of the GP model above the constraint threshold, to discourage such region from being explored. Furthermore, the same model behaves approximately as a standard GP around the three successful observations (black dots).
The objective cost $f$ is modeled with a standard GP, $f \sim \mathcal{GP}(0,k(x,x^\prime))$. Because no cost observation is obtained upon failure, the GP model in $f$ is not updated with such failures, as shown in \cref{fig:GPCR_all} (top).
The same approach has been recently adopted in \cite{englert2018learning}, in the context of robot learning.
 
The second difference is that in \eicc, the constraint threshold $c_j$ (cf. \cref{sec:controlprob}) is assumed unknown and estimated from data, while in BOC, $c_j$ is user-defined. In \cref{fig:GPCR_all} (middle), \gpcr~estimates  the constraint threshold $\hat{\thres}$ (dashed line) right above the successful observations.

The \eicc~framework uses the acquisition function \eqref{eq:eic}, which is illustrated in \cref{fig:GPCR_all} (bottom). As can be seen, the areas in the right, where the constraint is violated, are discouraged. After $M$ iterations, \eicc~reports the expected objective cost that satisfies the constraints with high probability \cite{gelbart2014bayesian}, computed as
\begin{equation}
x_\text{best} = \argmin_{x \in \dom} \mu(x)\;\; \text{s.t. } \Gamma(x) \geq 1 - \delta,
\label{eq:mincons}
\end{equation}
where $\delta \in [0,1]$ is typically set to a small number and $\Gamma$ is given in \eqref{eq:eic}. \cref{alg:eicc} summarizes \eicc~in pseudocode with a single constraint for simplicity, where $y_t^f$ and $y_t^g$ constitute noisy observations of the objective $f$ and the constraint $g$, respectively.

\begin{algorithm}[t!]
\caption{Expected improvement with crash constraints}
\begin{algorithmic}[1]
\State \textbf{Input:} Hyperpriors $p(\theta)$, $p(c)$, covariance function $k$, $M > 1$, $\tau$, $\delta$, $f \sim \GP{0}{k(x,x^\prime)}$, $g \sim \GPCR{0}{k(x,x^\prime)}$
\State \textbf{Initialization} Choose $x_1 \in \dom$ randomly
\For{$n=1\text{ to }M-1$}
	\State $\{ y^f_n,y^g_n,l_n \} \gets$ \Call{RobotExperiment}{$x_n$} 	\State $\mathcal{D}_n^f \gets \mathcal{D}_n^f \cup \{ y^f_n,x_n \}$
	\State $\mathcal{D}_n^g \gets \mathcal{D}_n^g \cup \{ y^g_n,l_n,x_n \}$
	\State Optimize hyperparameters $\hat{\theta}^f_n$ using MAP \Comment{\cite{Rasmussen2006Gaussian}}
	\State Optimize hyperparameters $(\hat{\theta}^g_n, \hat{c}^g_n )$ using MAP \Comment{\eqref{eq:c_opti}}
	\State $x_{n+1} = \arg\max_{x\in \dom} \alpha(x)$ \Comment{Next candidate via \eqref{eq:eic}}
\EndFor
\State Estimate constrained global minimum $x_\text{best}$ \Comment{\eqref{eq:mincons}}
\State \Return $x_\text{best}$
\Statex
\Function{RobotExperiment}{$x_n$}

\State Run robot experiment for $\tau$ time steps.
\If{Task failure}
	\State \Return $\{ \varnothing,\varnothing,l=\lu \} $
\Else
	\State \Return $\{ y^f,y^g,l=\ls \}$
\EndIf

\EndFunction
\end{algorithmic}
\label{alg:eicc}
\end{algorithm}

In the following, we explain the \gpcr~model that \eicc~uses to model the constraints and also how this model can be used to estimate the constraint threshold while learning. For ease of presentation, we consider therein a single constraint $g$.

 \section{Bayesian model for hybrid data and unknown constraint threshold}
\label{sec:gpcrsec}
The considered problem setting assumes that neither the objective $f$ nor the constraint $g$ can be observed upon failure (cf. \cref{sec:controlprob}). In such case, the only acquired data is a binary label indicating ``failure.'' In contrast, upon constraint satisfaction we obtain real observations of $f$ and $g$ \emph{and} a binary label indicating ``success.''
In the following, we propose a GP model for $g$ able to handle \emph{hybrid} data: binary labels and continuous values.

\subsection{Gaussian process for classified regression (\gpcr)}
\label{ssec:gpcr}

Let us assume that for each controller parametrization $\inx_i$ we simultaneously obtain two types of observations: (i) a binary label $l_i\in \left\lbrace \lu,\ls\right\rbrace$ that determines whether the experiment was failure or success, respectively, and (ii) a noisy constraint value $y_i \in \R$ only when the experiment is successful:
\begin{equation}
(y_i,l_i) = 
\left\lbrace
\begin{array}{lrl}
(g_i + \varepsilon, &\ls), & \text{ if } \inx_i \text{ is success} \\
(\varnothing,&\lu), & \text{ if } \inx_i \text{ is failure} \\
\end{array}
\right.
,
\label{eq:output}
\end{equation}
where we have used the shorthand notation $g_i = g(\inx_i)$, $y_i = y(\inx_i)$, and $l_i = l(x_i)$, and $\varepsilon \sim \N{\varepsilon}{0,\varnoise}$. The observation model can be represented with a probabilistic likelihood
\begin{equation}
p(y_i,l_i|g_i) = \left[ H(\thres-g_i)\N{y_i}{g_i,\varnoise} \right]^{l_i} \left[ H(g_i-\thres) \right]^{1-l_i},
\label{eq:lik}
\end{equation}
where
$H(z)$ is the Heaviside function, i.e., $H(z) = 1$, if $z \geq 0$, and $0$ otherwise. The likelihood \eqref{eq:lik} captures our knowledge about the latent constraint\footnote{The likelihood function \eqref{eq:lik} is an unnormalized density due to the presence of the Heaviside functions. Using unnormalized likelihood functions is common in the context of GP classification models \cite{bachoc2019gaussian,hernandez2014mind}.}: if $x_i$ is a failure ($l_i=0$), all we know about $g_i$ is that it takes any possible value above the threshold $\thres$, with all values $g_i\geq\thres$ being equally likely, but we never specify what that value is. To this end, 
the likelihood $p(y_i,l_i=0|g_i) = H(g_i-\thres)$ places all probability mass above the threshold $\thres$. On the contrary, if $x_i$ is successful ($l_i=1$), all the probability mass falls below $\thres$ for consistency, shaped as truncated Gaussian noise centered at $g_i$, i.e., $p(y_i,l_i=1|g_i)=H(\thres-g_i)\N{y_i}{g_i,\varnoise}$.

Let the latent constraint be observed at locations $X=\{X_\text{s},X_\text{u}\}$, which entails both, successful $X_\text{s}=\left\lbrace x_i\right\rbrace_{i=1}^{\Ns}$ and failing controllers $X_\text{u}=\left\lbrace x_i\right\rbrace_{i=\Ns+1}^N$. The corresponding latent constraint values at $X_\text{s}$ and  $X_\text{u}$ are $\fs=\left[g_1,g_2,\dots,g_{\Ns}\right]^\top$, and $\fu=\left[g_{\Ns+1},g_{\Ns+2},\dots,g_N\right]^\top$, respectively. The latent constraint values $\f = \left( \fs, \fu \right)$
induce observations grouped in the hybrid set of observations $\sobs = \left\lbrace y_i,l_i \right\rbrace_{i=1}^{\Ns} \cup \left\lbrace l_i \right\rbrace_{i={\Ns+1}}^N$, which contains both, discrete labels and real scalar values.
We assume such observations to be independent and identically distributed, which allows the likelihood to factorize $p(\sobs|\f) = \prod_{i=1}^N p(y_i,l_i|g_i)$. Then, the likelihood over the data set $\sobs$ becomes
\begin{equation}
p(\sobs|\f) = \prod_{i=1}^{\Ns} H(\thres-g_i)\mathcal{N}(y_i|g_i,\varnoise) \hspace{-7pt}  \prod_{i=\Ns+1}^N \hspace{-7pt}  H(g_i-\thres).
\label{eq:lik_elabo}
\end{equation}
The posterior follows from Bayes theorem $p(\f|\sobs) \propto p(\sobs|\f)p(\f)$, 
with zero-mean\footnote{A non-zero constant mean function could be included in the model with no additional complexity.}
Gaussian prior
$p(\f) = \mathcal{N}(\f|\bm{0},\bm{K})$, $\left[ \bm{K} \right]_{i,j} = k(x_i,x_j)$, and multivariate likelihood~\eqref{eq:lik_elabo}.
The terms in ~\eqref{eq:lik_elabo} can be permuted and the Gaussian factors can be grouped in a multivariate Gaussian $\prod_{i=1}^{\Ns} \mathcal{N}(y_i|g_i,\varnoise)=\mathcal{N}(\Ys|\fs,\varnoise \bm{I})$. The product of two multivariate Gaussians of different dimensionality $\mathcal{N}(\Ys|\fs,\varnoise \bm{I}) \mathcal{N}(\f|\bm{0},\bm{K})$ equals another unnormalized Gaussian $\kappa\N{\f}{\tilde{\bm{m}},\tilde{\bm{\Sigma}}}$, whose mean, variance and scaling factor depend on the observations and the noise, i.e.,
$\tilde{\bm{m}} = \tilde{\bm{m}}(\Ys,\varnoise)$,
$\tilde{\bm{\Sigma}} = \tilde{\bm{\Sigma}}(\varnoise)$,
and $\kappa = \kappa(\Ys,\varnoise)$.
Hence, the posterior becomes
\begin{equation}
p(\f|\sobs) \propto \N{\f}{\mt,\vart}\prod_{i=1}^{\Ns} H(\thres-g_i) \hspace{-7pt}  \prod_{i=\Ns+1}^N \hspace{-7pt}  H(g_i-\thres).
\label{eq:my_post}
\end{equation}
The Heaviside functions in \eqref{eq:my_post} restrict the support of $p(\f|\sobs)$ to an unbounded hyperrectangle with a single corner at location $g_i=c,\;\forall i \in \{1,\dots,N\}$. Because this particular shape complicates the computation of the predictive distribution, we approximate the posterior $p(\f|\sobs)$.
Among the many possible methods \cite{nickisch2008approximations}, we opted here for expectation propagation (EP), a variational approach that approximates well the right hand side of \eqref{eq:my_post} with a multivariate Gaussian $q(\f) = Z_\text{EP}\N{\f}{\meanpost,\varpost}$ \cite{cunningham2011gaussian}. In EP, the moments $\meanpost$, $\varpost$ and $Z_\text{EP} \in \R_{ > 0}$
are matched to those of the right hand side of \eqref{eq:my_post}. The zero-th order moment $Z_\text{EP}$ is a constant that approximates the model evidence $p(\sobs) = \int_{\f} p(\sobs | \f) p(\f) \df \simeq \int_{\f} q(\f) \df = Z_\text{EP}$, which will be further discussed in \cref{ssec:thres}.

Since the approximate posterior $q(\f) \simeq p(\f|\sobs)$ is Gaussian, the predictive distribution at some unobserved location $x$ is a univariate Gaussian $p(g(x)) = \N{g(x)}{\mu(x),\sigma^2(x)}$, whose moments are given analytically~\cite[Sec. 3.4.2]{Rasmussen2006Gaussian}
\begin{equation}
\hfilneg
\begin{array}{rl}
\mu(x) = & \bm{k}^\top(x)K^{-1}\meanpost \\ 
\sigma^2(x) = & k(x,x) - \bm{k}^\top(x)K^{-1} \left( \bm{I} -  \varpost K^{-1} \right) \bm{k}(x).
\end{array}
\hspace{1000pt minus 1fil}
\label{eq:pred_q_moments}
\end{equation}
The above equations are also valid for a set of unobserved locations $\bm{x} = \left[x_1,\ldots,x_T\right]$, in which case $p(g(\bm{x})) = \N{g(\bm{x})}{\mu(\bm{x}),\Sigma(\bm{x})}$ is the sought \gpcr~model, indexed at $\bm{x}$.
 Generally, we say that $g \sim \mathcal{GPCR}(0,k(x,x^\prime))$ is modeled with \gpcr, with zero-mean and kernel $k$. 

The predictive probability of constraint satisfaction at location $x$ is given by
$$\Prob{l(x) = +1} = \int_{g(x)} H(\thres-g(x)) p(g(x))\text{d}g(x),$$
where the dependency on $x$ has been introduced for clarity. Since $p(g(x))$ is a univariate Gaussian, such integral can be resolved analytically
\begin{equation}
\Prob{l(x) = +1} = \CDF{\tfrac{\thres - \mu(x)}{\sigma(x)}},
\label{eq:prob_sta}
\end{equation}
where $\mu(x)$ and $\sigma(x)$ are given in \eqref{eq:pred_q_moments}, and $\CDF{\cdot}$ is the cumulative density function of a standard normal distribution.

Both, the likelihood model \eqref{eq:lik_elabo} and the approximate posterior \eqref{eq:my_post}, depend on the threshold parameter $\thres$, which
can be seen as a discriminator that distinguishes instability from stability in the axis of the constraint value. 
While $\thres$ has been treated so far as a fixed value, we consider it now a hyperparameter. Next, we explain how it can be estimated.

\subsection{Constraint threshold as a hyperparameter}
\label{ssec:thres}
We treat fully Bayesian the model hyperparameters $\theta$ (e.g., lengthscales of the kernel) and the constraint threshold $c$ (cf. \cref{ssec:gpcr}) and seek to compute their posterior probability distribution $p(\thres,\theta | \sobs) \propto p(\thres)p(\theta) p(\sobs | \thres, \theta)$. While $p(c)$ and $p(\theta)$ are given, the model evidence $p(\sobs | \thres, \theta) = \int_{\f} p(\sobs|\f,\thres,\theta)p(\f) \df$ involves computing an intractable integral. Such integral can be approximated as $\int_{\f} q(\f;\thres,\theta) \df = Z_\text{EP}(\thres,\theta)$ \cite{cunningham2011gaussian} (cf. \cref{ssec:gpcr}), where 
the dependency on $\thres$ and $\theta$ has been introduced for clarity.
This enables many different approximations for $p(\thres,\theta | \sobs)$. Herein, we directly estimate $\thres$ and $\theta$ via \emph{maximum a posteriori} (MAP) \cite{Rasmussen2006Gaussian}, given by
\begin{equation}
\thresopt,\,\hat{\theta} = \argmax_{\thres > y_\text{max},\;\theta} \log Z_\text{EP}(\thres,\theta) + \log p(c) + \log p(\theta),
\label{eq:c_opti}
\end{equation}
where the prior over the constraint threshold is defined as $p(\thres) = \text{Gamma}(\thres - y_\text{max})$. We shift the origin of the distribution to restrict the support above the worst constraint value among the successful observations, i.e., $\thres >  y_\text{max}$, with $ y_\text{max} = \max \{ y_i \}_{i=1}^{\Ns}$.

\subsection{Constraint modeled with \gpcr~in \eicc}
The proposed \eicc~framework suggests candidate evaluations according to \eqref{eq:eic}, where the constraints are modeled with \gpcr. With $G \geq 1$ constraints, the probability of the constraints being satisfied is given as $\Gamma (x) = \prod_{j=1}^G \Prob{l_j(x) = \ls}$, where $\Prob{l_j(x) = \ls}$ is computed as in \eqref{eq:prob_sta}. For simplicity, we consider in the following section $G=1$. However, \eicc~is generally applicable when $G \geq 1$, as its acquisition function~\eqref{eq:eic} (EIC) has been shown to work with multiple constraints \cite{hernandez2016general} and it only requires the constraint to be modeled with a GP.

 \section{Results}
\label{sec:results}

In this section, we assess the performance of \eicc~in two constrained optimization scenarios where no data is obtained upon constraint violation: numerical benchmarks and a real robot platform. The overall goal is to show that expert knowledge can be reduced for solving the problem of crash constraints by (i) comparing the performance of \eicc~against existing heuristic strategies that penalize failed experiments with user-defined costs and (ii) showing that the constraint threshold can be learned from real experimental data.
 In the following, we describe the experimental settings common to both scenarios and then describe the experiments and results of each scenario.

\subsection{Overall experimental setting}
We quantify the learning performance using \emph{simple regret}, which is a common performance metric in BO \cite{wang2017mes,hernandez2016general} and is defined as $r_n = y(x_n) - \min_{x \in \dom} f(x)$.
The input domain is normalized to the unit hypercube $\dom = \left[0,1 \right]^D$. 

For all GPs we assume a zero-mean prior and use Mat\'{e}rn kernel $\nicefrac{5}{2}$ \cite[Sec. 4.2]{Rasmussen2006Gaussian} with scaling factor $\kappa > 0$ and lengthscale $\lambda > 0$. A prior probability distribution is assumed over the lengthscales and variance of the kernel, and over the constraint threshold $c$ for the \gpcr~model. The observation noise for both $f$ and $g$ is modeled as in \cref{ssec:GPs} and \eqref{eq:output}, respectively, with fixed noise variance $\varnoise$.
The
hyperparameters $\theta=\{\lambda, \kappa\}$ are re-estimated at each iteration using maximum a posteriori (MAP), which in the case of the \gpcr~model is computed as in \eqref{eq:c_opti}. Further details about the hyperparameters can be found in our Python implementation\footnote{\url{https://github.com/alonrot/classified_regression}} of \eicc~and \gpcr.

\subsection{Comparing against BO with heuristic penalties and BOC}
\label{ssec:heur}

\begin{figure}[t!]
\centering
\includegraphics[width=\columnwidth]{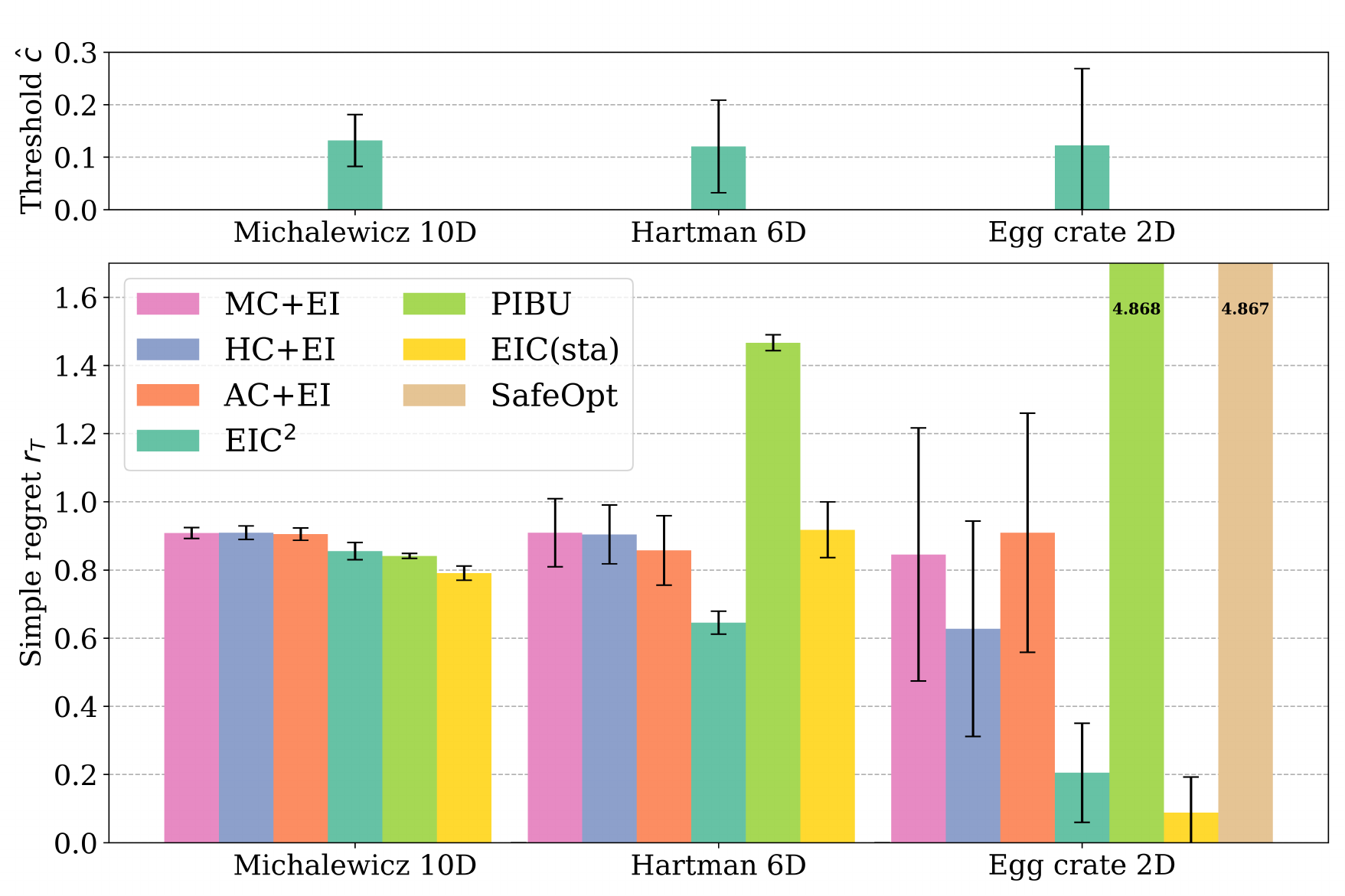}
\caption{Top: Estimated averaged constraint threshold $\hat{c}$ computed via \eqref{eq:c_opti} within the \gpcr~model used in \eicc. The black error bars represent one standard deviation. Bottom: For each benchmark, the simple regret at the last iteration is reported for each heuristics and \eicc. The black error bars represent half a standard deviation.}
\label{fig:regrets_all}
\end{figure}

In related work on learning controller parameters with BO \cite{yuan2019bayesian,marco2016automatic,calandra2016bayesian,rai2018bayesian,antonova2016sample}, robot failures are not interpreted as constraint violation. Instead, such works solve an unconstrained optimization problem using BO and simply assign a high cost to failure cases, to emphasize the poor performance of such experiments. However, such high cost is heuristically chosen and typically requires expert knowledge.
Herein, we compare, in numerical benchmarks, \eicc~against three plausible heuristic strategies to penalize failures: using a high cost (HC) \cite{marco2016automatic,yuan2019bayesian,rai2018bayesian,antonova2016sample}, using a middle cost (MC), and using an adaptive cost (AC). In the first, the penalizing cost is fixed before starting the learning experiments and is chosen as an upper bound on the function to optimize\footnote{Such upper bound is accessible in numerical benchmarks, but generally unknown when learning on real systems.}.
Similarly, in the second, the penalizing cost is chosen as the value of the initial evaluation. Finally, in the third, the penalizing cost of all failed experiments is re-adapted at each iteration to be the maximum (worst) observed cost value so far. As in \cite{calandra2016bayesian,rai2018bayesian,antonova2016sample}, the unconstrained BO algorithm used for each strategies is EI (cf. \cref{sssec:ei}). The purpose of these comparisons is to emphasize that with \eicc~no expert knowledge is required to penalize failures, and that the constraint threshold can be learned from data.

We also compare \eicc~against three BOC methods: the well-known EIC (cf. \cref{sssec:eic}) as a baseline, and two popular methods with applications in robotics, \pibu\footnote{\url{https://github.com/etpr/con_bopt}}~\cite{englert2018learning,driess2017constrained} and \safeopt~\cite{berkenkamp2016safe}, whose implementations are available online. Contrary to \eicc, EIC and \safeopt~model the constraint using a standard GP and need to be informed about the value of the constraint threshold $c$. \pibu~assumes binary constraint observations, hence, it models the constraint using a GP classifier.

\subsubsection{Experimental choices}

We use three challenging benchmarks for global minimization that exhibit multiple local minima: Michalewicz 10-dimensional function, Hartman 6-dimensional, and Egg Crate two-dimensional function \cite{survey2013benchmarks}, which are common benchmarks in BO, also used in \cite{wang2017mes}.

The aforementioned cost objectives are constrained to
a scalar function $g(x) = \prod_{d=1}^D \sin \left( 2\pi x_d \right)$, which divides the volume in $2^D$ sub-hypercubes, among which there are $2^{D-1}$ safe sub-hypercubes, alternated with the unsafe ones. Constraint satisfaction is determined by $g(x) \leq c$, with $c = 0$. To compute the regret $r_n$ we use the unconstrained global minimum value of each function, reported in \cite{survey2013benchmarks}.

We run \cref{alg:eicc} and the other methods for $M=100$ iterations and assess consistency by repeating it 100 times. In all cases, the initial point is randomly sampled within the safe regions for simplicity.

\subsubsection{Results}
In \cref{fig:regrets_all} (bottom), we see that \eicc~consistently achieves lower average regret than the aforementioned heuristics, which confirms that (i) treating failures apart by modeling them using a constraint is more beneficial than heuristically penalizing failures and (ii) not knowing the constraint threshold a priori does not affect significantly the learning performance of \eicc. Moreover, \eicc~outperforms the other methods in the 6D case. Unsurprisingly, in the other cases, EIC exhibits lower regret than \eicc, as EIC is informed about the true constraint threshold ($c=0$), while \eicc~learns it. While \pibu~exhibits similar regret in the 10D case, it performs worse in 6D and 2D. The reason is that \pibu~explores locally around the initial point in order to expand the safe area, hence, missing alternative safe areas of lower regret. Similarly, \safeopt~never leaves the initial safe region. Because \safeopt~scales poorly with dimensionality, we could not compute results for 6D and 10D.

\begin{table}[t!]
	\setlength{\tabcolsep}{8pt}
	\centering
	\caption{Percentage of safe evaluations. Average and standard deviation $(\cdot)$ over 100 experiments.}
	\begin{tabularx}{\columnwidth}{c|c|c|c}
		
					& 	Michalewicz 10D 	& 	Hartman 6D 		& 	Egg crate 2D \\
					\hline
		MC + EI		& 	$50.48$ $(5.28)$		&	$63.54$ $(7.65)$	&	$70.38$ $(5.87)$ \\
		HC + EI		& 	$53.02$ $(5.59)$		&	$64.53$ $(6.92)$	&	$69.98$ $(4.86)$  \\
		AC + EI		& 	$51.39$ $(5.11)$		&	$64.16$ $(7.20)$	&	$69.19$ $(6.19)$  \\
		\eicc		& 	$77.90$ $(7.78)$		&	$69.40$ $(5.44)$	&	$79.41$ $(12.98)$  \\
		PIBU		& 	$86.13$ $(0.97)$		&	$80.86$ $(1.95)$	&	$96.12$ $(1.38)$  \\
		EIC			& 	$92.92$ $(7.93)$		&	$48.14$ $(15.71)$	&	$75.12$ $(4.66)$  \\
		SafeOpt		& 		  $-$				&         $-$       	&   $100.00$ $(0.00)$ \\

	\end{tabularx}
	\label{tab:safe_evals}
\end{table}

\cref{tab:safe_evals} shows that \eicc~finishes the exploration with a higher number of safe evaluations on average than the BO heuristics, as it leverages the information of \gpcr~to reason about unpromising areas. Also, \eicc~finds the best trade-off between failure avoidance and performance while not being informed about the constraint threshold.
The baseline, EIC, reaches a higher number of safe evaluations in 10D, and a similar number in 2D, while showing the lowest regret in 2D. However, EIC plays with two advantages with respect to \eicc: (i) it is informed about the true constraint threshold and (ii) upon failure, it receives the true constraint observation.

In \cref{fig:regrets_all} (top), the constraint threshold $\hat{c}$ is estimated consistently near the true threshold ($c=0$), i.e., a $\sim$\SI{5}{\percent} of the total range ($[-1,+1]$). On average, the $\hat{c}$ values are reported slightly above zero because the Gamma hyperprior $p(c)$ places the mean of the distribution slightly above $y_\text{max}$ (cf. \cref{ssec:thres}).

\subsection{Experiments on a real jumping quadruped}
\label{ssec:jumping}
Herein, we describe and analyze a set of experiments in which a quadruped learns to jump as high as possible. The higher the jump, the higher is the current needed to achieve it. 
Excessive demands of current cause a voltage drop that triggers a safe mode on the control boards. When this happens, all the motors shut down, which results in a failed jump, in which the robot lands improperly and crashes.
The purpose of these experiments are two fold: (i) illustrate that \eicc~exploits failures on its benefit to steer the search towards promising areas, although no data is obtained upon failure, and (ii) show that \eicc~allows to estimate the maximum current above which the robot will fail.

\subsubsection{Experimental setting}
For the learning experiments we use Solo \cite{grimminger2020solo} (cf. \cref{fig:eye_catcher}), a lightweight quadruped with two degrees of freedom per leg that uses high-torque brushless DC motors, which enable very high vertical jumps. A kinodynamic planner \cite{ponton2020efficient} uses an approximate robot model to compute offline the needed feedforward torques and desired state trajectories for a high jump.
To account for modeling errors, we use a PD controller with variable gains.
Such gains change smoothly from an initial to a final value, both of which need to be tuned in order to achieve the desired jump. Poor tuning results in poor tracking and extremely low jumps. Hence, we use \eicc~to automatically tune the gains toward better tracking and higher jumps. By using symmetry properties in the platform and the task, we can reduce the dimensionality of the tuning problem. To this end, we couple the P and D gains from all the knees into a single set of gains, and same for all the hips.
In addition, we fix the final value of the P and D variable gains, which leaves four parameters to tune: $x = \left( \mathrm{P}_\text{init}^\text{knee}, \mathrm{P}_\text{init}^\text{hip}, \mathrm{D}_\text{init}^\text{knee}, \mathrm{D}_\text{init}^\text{hip} \right)^\top$. The parameters are normalized to lie within the unit hypercube $x \in [0,1]^4$.

For a certain controller parametrization that results in no failure,
the cost objective 
is defined as
$f(x) = -\max \{ h_1,\ldots,h_\tau \}$,
where $h_k$ is the height of the jump in meters measured at time step $k$ and $\tau$ is the number of time steps elapsed until landing. Each $h_k$ is measured at the center of the robot base using a motion capture system which operates at \SI{200}{\hertz}. The quadruped initiates all jumps from the same resting position, with the legs slightly flexed. In this position the robot is \SI{0.25}{\meter} high. 
The constraint is defined as
$g(x) = \sum_{m=1}^8 i_m$,
where $i_m$ is the peak current measured in motor $m$ during the jump. Constraint satisfaction is defined as $g(x) \leq c$, where $c$ represents the maximum sum of currents, above which the robot will shut down and crash. The constraint threshold $c$ is unknown a priori and can only be revealed with empirical evidence\footnote{Alternatively, a thorough analysis of the limitations of the DC motors given by the manufacturer could lead to a theoretical approximation of this value. However, this alternative typically requires expert knowledge, which might not always be available.}.
Critically, an underestimated threshold would wrongly reduce the space of successful controllers while an overestimated threshold would wrongly classify failure regions as successful. Instead, we let \eicc~estimate the constraint threshold $c$ using the \gpcr~that models $g$.

The noise variances are measured by repeating ten times an initial safe jump and computing the empirical variance of the height and the current.

\begin{figure}[t!]
\centering
\includegraphics[width=\columnwidth]{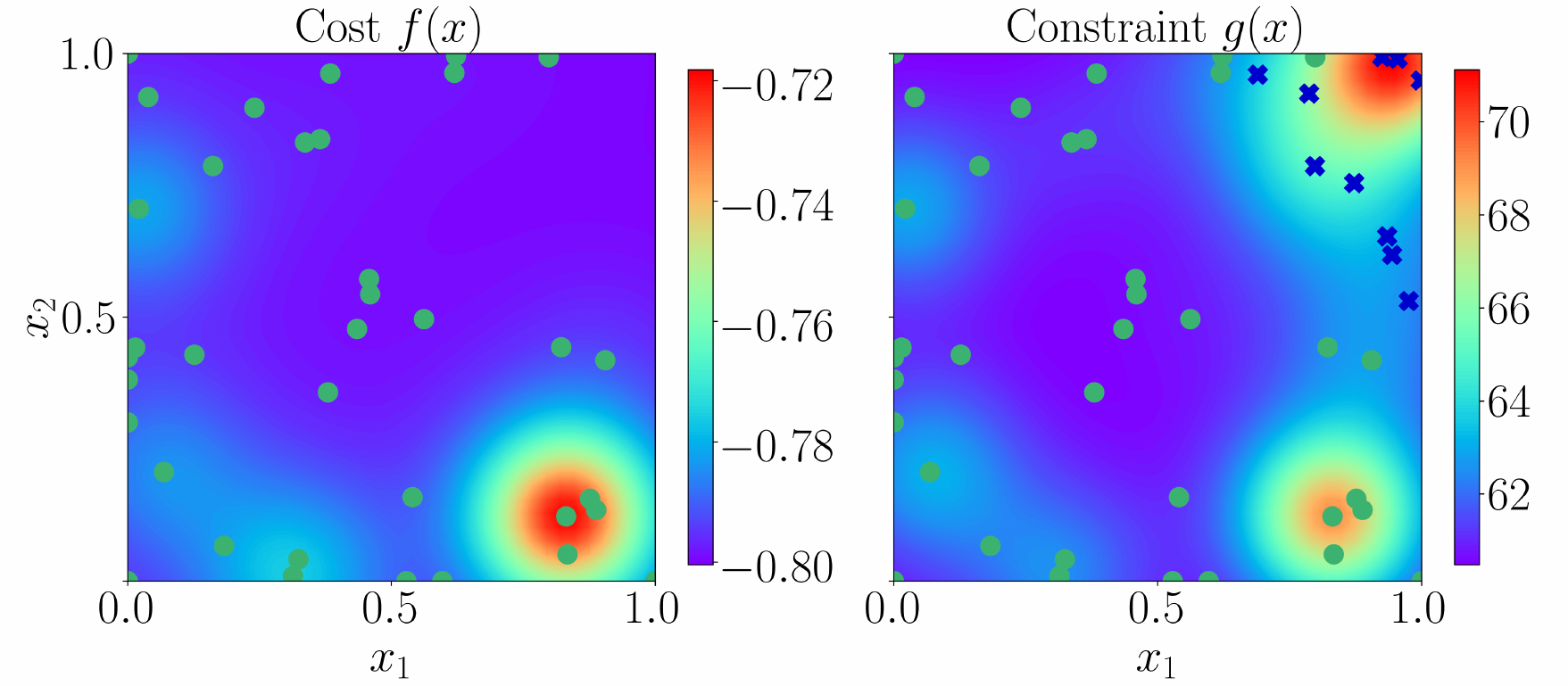}
\caption{Two-dimensional slice of the GP posterior mean, computed by fixing the third and fourth coordinates of the input vector, i.e, $x=(x_1,x_2,0.7,0.4)$. Successful evaluations (green dots) are shown in both plots, while failures (blue crosses) are shown only in the left plot. This is to emphasize that the GP that models $f$ is not updated with the failure data points, while \gpcr~is updated using both: the successful observations and the failures.}
\label{sfig:quadru:slice}
\end{figure}

We start the experiments at a corner of the domain, i.e., $x=(0,0,1,1)$ and run them for $M=50$ iterations, after which we report the best successful observation obtained.

\subsubsection{Results}

After the exploration, \eicc~succeeded to report a remarkably high jump. Importantly, the \gpcr~that models $g$ was able to handle failures cases in which no current measurement was obtained, together with successful observations, in which the current is measured. In addition, \gpcr~was able to estimate the constraint threshold.

As shown in \cref{sfig:quadru:slice}, most successful jumps were encountered during the search by \eicc~in the upper half of the two-dimensional slice. However, 10 failures were encountered at the top right corner, as a consequence of high P gains.
Between iterations 12 and 34, \eicc~reported mostly high jumps mixed in with the failures. Because of this, it is likely that the constrained minima lies close to the constraint boundary. Consequently, underestimated user-defined thresholds could have discouraged such constraint boundary from being thoroughly explored, hence potentially missing high jumps. After iteration 34, \eicc~led the search away from the region of failures and and targeted safer areas where jumps were generally not that high.

The controller parametrization that provided the highest jump was found at $x_\text{highest} = (0.62, 1.0, 0.44, 0.58)$ (iteration 26), which is close to the area of failures. After the search, we double-checked $x_\text{highest}$ by executing ten jumps, which resulted in a current of $104.48 \pm 1.32\;\mathrm{A}$ and a height of $0.784 \pm 0.0023\;\mathrm{m}$, which is about $\times 3$ times higher than its resting position. The complementary video\footnote{\url{https://youtu.be/RAiIo0l6_rE}}
summarizes the learning process and shows the learned jump. In \cite{grimminger2020solo}, significant manual tuning was required to achieve high jumps on the same robot. However, their highest jump was reported at \SI{0.65}{\metre}, while our framework automates the tuning process and gains about \SI{13}{\centi\metre} (about \SI{34}{\percent} improvement). We also tested the estimated constrained global minimum $x_\text{best}$ by solving $\eqref{eq:mincons}$ with $\delta=0.1$. However, the performance was slightly worse than the one reported with $x_\text{highest}$.

The constraint threshold was reported by \gpcr~at $\hat{c} = \SI[per-mode=symbol]{112.34}{\ampere}$. This value was unknown and unaccessible before the experiments, but learned as a hyperparameter of \gpcr. This value can be stored for later implementations of safety mechanisms on the robot, e.g., for further learning experiments or demonstrations.

 \section{Conclusions}

In this paper, we have addressed the problem of learning with crash constraints, in which no data is obtained upon failure. To this end, we have proposed \eicc, a constrained Bayesian optimization framework that models the constraint with \gpcr, a novel GP model able to handle discrete and continuous observations. We demonstrated the effectiveness of \eicc~in numerical benchmarks up to ten dimensions and on a real jumping quadruped, where the learned best jump improved by \SI{34}{\percent} over the results obtained with manual tuning.
Also, \gpcr~learned the constraint threshold, unknown a priori, which mitigates the need of expert knowledge to determine it. 
In future work, we will perform further robot experiments in more challenging and higher dimensional robot platforms with multiple constraints and compare \eicc~against multi-output Gaussian process models.

\section*{Acknowledgment}
The authors thank Felix Grimminger, for his support with the hardware during the experiments, Manuel W\"{u}thrich for insightful discussions, and Maximilien Naveau for facilitating software infrastructure. 
\bibliographystyle{IEEEtran}
\bibliography{root}

\end{document}